\documentclass[letterpaper, 10 pt, conference]{ieeeconf}  % Comment this line out if you need a4paper

\IEEEoverridecommandlockouts                              % This command is only needed if 
\usepackage{cite}
\usepackage{graphics}
\usepackage{epsfig}
\usepackage{amsmath}
\usepackage{booktabs}
\usepackage{amsmath,amssymb,amsfonts}
\usepackage[ruled,linesnumbered,commentsnumbered,vlined]{algorithm2e}

\title{\LARGE \bf
ESCoT: An Enhanced Step-based Coordinate Trajectory Planning Method for Multiple Car-like Robots
}

\author{Junkai Jiang$^{1}$, Yihe Chen$^{1}$, Yibin Yang$^{1}$, Ruochen Li$^{1}$, Shaobing Xu$^{1}$, and Jianqiang Wang$^{1*}$% <-this % stops a space
\thanks{This work was supported by National Natural Science Foundation of China, Science Fund for Creative Research Groups (Grant No. 52221005) and National Natural Science Foundation of China (Grant No. 52131201).}% <-this % stops a space
\thanks{*Corresponding author: Jianqiang Wang.}% <-this % stops a space
\thanks{$^{1}$The authors are with the School of Vehicle and Mobility, Tsinghua University, Beijing, China.
        {\tt\small \{jiangjk21, chenyihe21, yyb19, lrc23\}@mails.tsinghua.edu.cn, \{shaobxu, wjqlws\}@tsinghua.edu.cn}}%
}

\begin{document}

\maketitle
\thispagestyle{empty}
\pagestyle{empty}

%%%%%%%%%%%%%%%%%%%%%%%%%%%%%%%%%%%%%%%%%%%%%%%%%%%%%%%%%%%%%%%%%%%%%%%%%%%%%%%%
\begin{abstract}

Multi-vehicle trajectory planning (MVTP) is one of the key challenges in multi-robot systems (MRSs) and has broad applications across various fields. This paper presents ESCoT, an enhanced step-based coordinate trajectory planning method for multiple car-like robots. ESCoT incorporates two key strategies: collaborative planning for local robot groups and replanning for duplicate configurations. These strategies effectively enhance the performance of step-based MVTP methods. Through extensive experiments, we show that ESCoT 1) in sparse scenarios, significantly improves solution quality compared to baseline step-based method, achieving up to 70\% improvement in typical conflict scenarios and 34\% in randomly generated scenarios, while maintaining high solving efficiency; and 2) in dense scenarios, outperforms all baseline methods, maintains a success rate of over 50\% even in the most challenging configurations. The results demonstrate that ESCoT effectively solves MVTP, further extending the capabilities of step-based methods. Finally, practical robot tests validate the algorithm's applicability in real-world scenarios.

\end{abstract}

%%%%%%%%%%%%%%%%%%%%%%%%%%%%%%%%%%%%%%%%%%%%%%%%%%%%%%%%%%%%%%%%%%%%%%%%%%%%%%%%
\section{Introduction}

Research on multi-robot systems (MRSs) has attracted growing attention in recent years, for its capability of accomplishing complicated missions more effectively and efficiently \cite{gautam2012review,dorri2018multi}. MRSs involve many critical problems, one of which is multi-agent path finding (MAPF). MAPF focuses on finding conflict-free paths for multiple agents in a shared environment, ensuring that each agent can navigate from its start position to its goal without colliding with others \cite{stern2019multi}. For the MAPF problem, two key assumptions are typically made: 1) the discretization of space, where agents occupy fixed grid locations, and 2) the discretization of agent action, where each agent can either wait or move one step in one of the four cardinal directions. These assumptions, however, make MAPF methods cannot directly applied to a large number of scenarios involving car-like robots, as the path planning is performed in continuous space and constrained by the kinematics of these robots. Therefore, considering the kinematic constraints of car-like robots and the spatial continuity, the multi-vehicle trajectory planning (MVTP) problem has become a critical issue in MRSs, and is the focus of this paper. 

\begin{figure}[htbp]
	\centering
	\includegraphics[width=6.5cm]{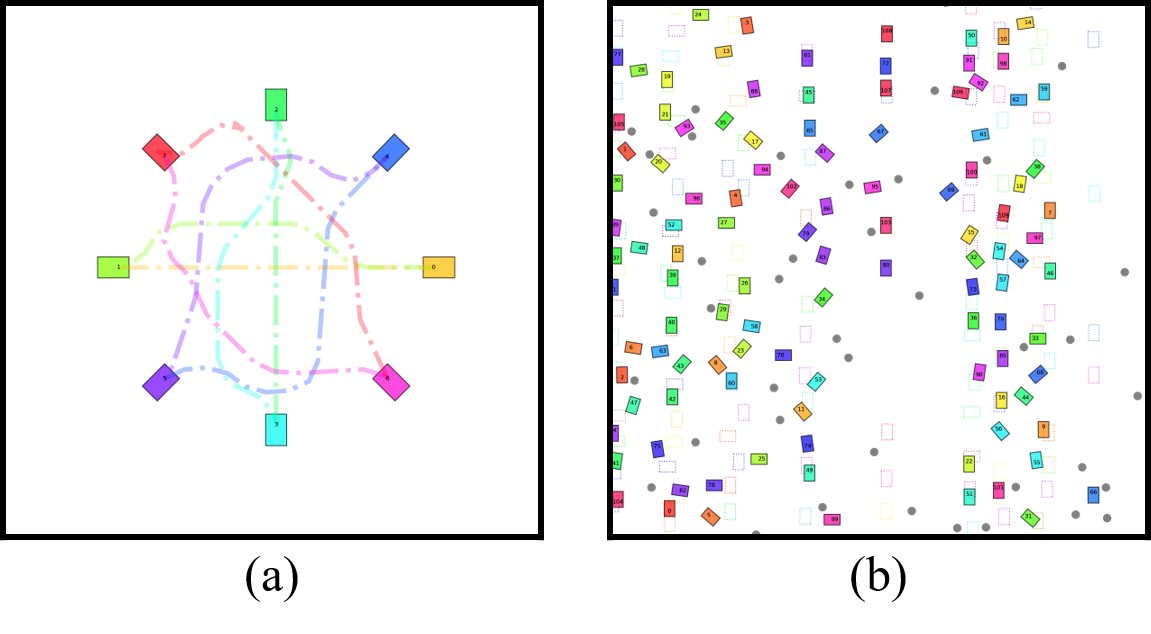}
	\caption{Two examples of MVTP problem. (a) shows a scenario with a sparse distribution of robots, where 8 robots are placed on a $50m\times50m$ map with no obstacles. The dashed lines represent the collision-free trajectories of each robot. While (b) illustrates a scenario with a dense distribution of robots, where 110 robots are located on a $100m\times100m$ map, along with 40 randomly distributed obstacles. The dashed boxes represent the goal positions of each robot.}  
	\label{fig:mvtp-example}
\end{figure}

MVTP aims to generate a series of collision-free trajectories for multiple vehicles (car-like robots), guiding them from their current positions to predefined goals in a known, unstructured environment, while minimizing travel time \cite{li2021optimal}.  MVTP has wide-ranging applications in various fields, such as warehouse automation \cite{yang2024csdo}, cooperative parking \cite{wu2019optimal}, and intersection management \cite{xu2024multi}. Fig.~\ref{fig:mvtp-example} presents two examples of MVTP problem.

MVTP problem has attracted significant research interest in the fields of robotics and artificial intelligence. Based on the difference in planning structure, existing studies can generally be divided into two categories: \textit{centralized methods}, where a central controller plans the trajectories for all robots, and \textit{distributed methods}, where each robot plan its trajectory independently based on environmental information. 

Centralized methods can further be classified into several categories, including grid search-based methods, sampling-based methods, constraint reduction methods, and coupled methods. Grid search is a typical approach in MAPF, and the grid search-based method for MVTP adapts its discretization strategy by discretizing the robot's poses and actions. Then the trajectories for each robot are generated through a search process \cite{yang2024csdo,wen2022cl,li2020efficient}. The efficiency and quality of these methods are directly related to the size of the search step. Moreover, the adaptation of high-efficiency MAPF methods to MVTP requires further exploration. Sampling-based methods typically extend path planning approachs for single car-like robots, such as probabilistic roadmap (PRM) and rapidly-exploring random trees (RRT), to MRSs \cite{lukyanenko2023probabilistic,shome2020drrt}. The quality and efficiency of sampling-based methods are influenced by the number of samples, requiring careful management of the trade-off between them. Constraint reduction methods aim to simplify the problem by progressively relaxing or reducing certain constraints during the planning process \cite{chen2015decoupled,ouyang2022fast}. This allows for more efficient path planning by reducing the complexity of the problem at each step. However, using constraint reduction methods to solve MVTP problem typically involves solving nonlinear programming (NLP) problems, which is challenging and time-consuming. The coupled method is based on a simple concept of treating all robots as a single high-dimensional agent and planning for this agent as a whole \cite{li2017centralized}. This approach offers theoretical completeness and optimality, but it can only be applied when the number of robots is small. As the number of robots increases, the branching factor grows exponentially, causing the search effort to increase rapidly and making it difficult to obtain solutions within a limited time frame.

Distributed methods solve MVTP in a manner similar to single-robot planning. In this approach, either robots are assigned predefined priorities, with later-planning robots treating earlier-planning robots as dynamic obstacles \cite{luis2020online,ma2023decentralized}, or agents resolve conflicts through local communication and cooperation \cite{ferranti2022distributed}. These methods are typically suitable for scenarios where robots and obstacles are sparsely distributed, as there are fewer conflicts to resolve, leading to high efficiency. However, in dense scenarios, distributed methods are prone to getting stuck in local optima due to the lack of coordination, resulting in a loss of both solution quality and efficiency.

Indeed, the main challenge of the MVTP problem lies in how to obtain the solution efficiently and effectively, and existing methods often struggle to achieve satisfactory performance in both solution quality and efficiency. Recently, step-based methods have emerged for MAPF, where at each step, the actions of all robots for the next time step are generated until all robots reach their goals \cite{okumura2022priority,okumura2023lacam,okumura2023improving}. These methods offer high computational efficiency, with solution quality gradually improving. Current research has applied the step-based approach to solve the MVTP problem, significantly enhancing solution success rates and time efficiency compared to other methods \cite{guo2024decentralized}. However, further improvements are still needed. On one hand, success rates remain insufficient in dense scenarios, and on the other hand, solution quality is still relatively low, leaving considerable room for optimization.

Therefore, the goal of this paper is to improve the performance of step-based methods for MVTP. We propose a new algorithm, Enhanced Step-based Coordinate Trajectory Planning (ESCoT) for multiple car-like robots, which is the main contribution of this work. Specifically, we introduce two strategies: collaborative planning for local robot groups and replanning for duplicate configurations. With these two strategies, ESCoT achieves a significant improvement in solution quality in sparse scenarios, as well as a further increase in solution success rates in dense scenarios. Furthermore, we conduct practical robot tests that verify the applicability of the algorithm in real-world robotic scenarios.

The rest of this paper is organized as follows. Section~\ref{Section 2} gives the definition of MVTP problem. Section~\ref{Section 3} introduces the details of the ESCoT algorithm and two key strategies. Experiments and validations are demonstrated in Section~\ref{Section 4}. Section~\ref{Section 5} provides the conclusions.

\section{Problem Definition}
\label{Section 2}

The MVTP problem can be defined by a ten-element tuple $\langle \mathcal{R},\mathcal{W},\mathcal{O},\mathcal{S},\mathcal{G},z,a,\mathcal{T}, \mathcal{P}, u \rangle$. $\mathcal{R}=\{r^1, \dots, r^m\}$ is the set of $m$ car-like robots. $\mathcal{W}$ is the workspace, composed of a map with dimensions $W \times H$ and $n$ static obstacles. The workspace occupied by the obstacles is denoted by $\mathcal{O}$. $\mathcal{S}=\{s^1, \dots, s^m\}$ and $\mathcal{G}=\{g^1, \dots, g^m\}$ represents the initial and goal configurations of each robot. $z=[x, y, \theta, \phi]^T$ is the state of a robot, where $x$ and $y$ are position coordinates, $\theta$ is the yaw orientation, and $\phi$ is the front wheel steering angle. The car-like robot is modelled as a rectangle with length $l$, width $w$, and a wheelbase of $l_b$. $a$ refers to the Ackermann-steering kinematics model. The workspace occupied by a robot's body with a given state $z$ is denoted by $\mathcal{T}(z)$, which is a subset of $\mathcal{W}$. The solution to the MVTP problem is a set of trajectories for all robots, symbolized as $\mathcal{P}=\{p^1, \dots, p^m\}$. $u=[v, \omega]^T$ represents the control input, where $v$ is the velocity and $\omega$ is the derivative (or differential) of $\phi$.

To achieve trajectory planning for multiple robots, we discretize time into intervals of $\Delta t$. In the following, for each variable, we use the superscript $i$ to denote that it is related to robot $r^i$, and the subscript $t$ to denote that it is related to the time step $t$.

The optimization objective of the MVTP problem is to minimize the task completion time, commonly referred to as \textit{makespan}. This is defined as the time when the last robot reaches its goal state. In this paper, for the sake of simplicity, we define $\Delta t$ as the unit time step, so the optimization objective can also be viewed as the \textit{makespan step}. The time at which robot $r^i$ reaches its goal state and remains stationary is denoted as $T^i$. Therefore, the optimization objective $T$ can be expressed as:
\begin{equation}
    T = \max_{i \in \{1,2,\dots,m\}} T^{i}
    \label{eqn:objective}
\end{equation}

As for the constraints, first, the MVTP must satisfy the boundary constraints of the robots' states:
\begin{equation} 
    \label{eqn:boundary_cons}
    z^{i}_0 = s^{i}, z^{i}_{T^{i}} = g^{i}, \forall i \in \{1,2,\dots,m\}
\end{equation}
Additionally, the robots must satisfy the collision-free constraints, which include collisions with obstacles and collisions between robots. These two conditions can be expressed as follows:
\begin{equation} 
    \label{eqn:static_cons}
    \mathcal{T}(z^{i}_{t}) \cap \mathcal{O}  = \emptyset, \forall t \geq 0, \forall i\in \{1,2,\dots,m\}
\end{equation}
\begin{equation} 
    \label{eqn:inter_cons}
    \mathcal{T}(z^{i}_{t}) \cap \mathcal{T}(z^{j}_{t}) = \emptyset, \forall t \geq 0, \forall i,j\in \{1,2,\dots,m\}, i \neq j
\end{equation}
Furthermore, the trajectories of the robots must adhere to the Ackermann-steering kinematics model and satisfy the kinematic constraints imposed by the robots' own characteristics, i.e.,
\begin{equation}
    \label{eqn:kine_cons}
    \begin{aligned}
    z_{t+1} &= z_{t} + \begin{bmatrix}
        v_t \cos \theta_t \\
        v_t \sin \theta_t \\
        v_t \tan(\phi_t)/L \\
        \omega \\
     \end{bmatrix} \Delta t, 
     \forall 0 \leq t < T
    \end{aligned}
\end{equation}
    
\begin{equation} 
    \label{eqn:control_max_cons}
    \left | v_t \right | \leq v_{\text{max}}, \left | \omega_t \right | \leq \omega_{\text{max}}, 
    \forall 0 \leq  t < T,
\end{equation}
\begin{equation} 
    \label{eqn:phi_max_cons}
    \left | \phi_t \right | \leq \phi_{\text{max}},  
    \forall 0 \leq  t \leq T
\end{equation}

Considering the optimization objective and all constraints, the MVTP problem can be formulated as an optimal control problem (OCP):

\begin{equation}
    \label{eq:ocp}
    \begin{aligned}
    \min & \quad \quad \textrm{(\ref{eqn:objective})}\\
    \textrm{s.t.} & \quad \textrm{(\ref{eqn:boundary_cons}),(\ref{eqn:static_cons}),(\ref{eqn:inter_cons}),(\ref{eqn:kine_cons}),(\ref{eqn:control_max_cons}),(\ref{eqn:phi_max_cons})}
    \end{aligned}
\end{equation}

The solution trajectories is recorded in $\mathcal{P}=\{p^1, \dots, p^m\}$, where $p^{i} = [z_0^{i}, z_1^{i}, ..., z_{T^{i}}^{i}..., z_{T}^{i}]$.

\section{Method}
\label{Section 3}

In this section, we first review the basic step-based method priority-inherited backtracking for car-like robots (PBCR) \cite{guo2024decentralized}, which has evolved from the priority inheritance with backtracking (PIBT) \cite{okumura2022priority} for MAPF. We then provide a detailed description of the proposed ESCoT algorithm, followed by the explanation on its two key strategies: collaborative planning for local robot groups and replanning for duplicate configurations. Finally, we discuss some implementation details of the algorithm.

\subsection{Review of PBCR} 
\label{pbcr-review}
\begin{figure}[htbp]
	\centering
	\includegraphics[width=8.5cm]{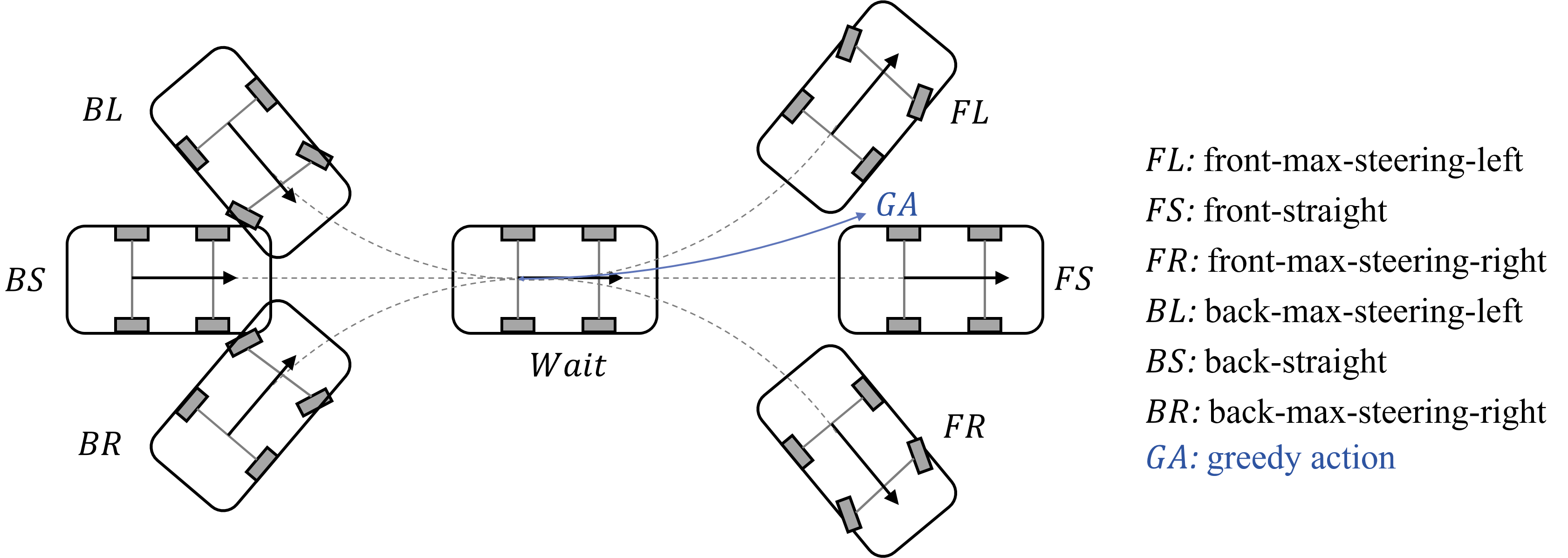}
	\caption{The action discretization to represent the kinematic model, which has at most eight actions.}  
	\label{fig:action-dis}
\end{figure}

To make the MVTP problem in continuous space feasible, PBCR and many other MVTP methods \cite{yang2024csdo,wen2022cl,li2020efficient} employ an action discretization approach to represent the kinematic model $a$, as shown in Fig.~\ref{fig:action-dis}. Except for the \textit{Wait} action, the step size for the other six basic actions is $v \cdot \Delta t$. In addition to this, PBCR introduces a greedy action (GA) that directly connects the robot's current state $z$ to the goal state $g$ using the hybrid A* \cite{dolgov2010path}, truncating the first segment of length $v \cdot \Delta t$. PBCR encourages robots to prioritize the \textit{GA} action, which improves the quality of the algorithm's results to some extent.

The complete process of the PBCR algorithm is summarized in Algorithm~\ref{algo:pbcr}. This is a typical step-based method, where as long as there are robots that have not reached their goal states, the next step for all robots is planned (lines 2-9). In each step of the planning, PBCR first establishes the priorities of all robots. Then, it plans for the robots sequentially from high to low priority. If the next step state of a high-priority robot conflicts with the current state of other robots, PBCR will perform priority inheritance and backtracking, which is the specific content of the PBCR function.

\begin{algorithm}[tp]
    \caption{The PBCR Algorithm}
    \label{algo:pbcr}
    \SetKwInOut{Input}{Input}
    \SetKwInOut{Output}{Output}

    \Input{$\mathcal{R},\mathcal{W},\mathcal{O},\mathcal{S},\mathcal{G}$}
    \Output{$\mathcal{P}=\{p^1, \dots, p^m\}$}

    $z^i_0 \gets s^i$: for each robot $r^i \in \mathcal{R}$ \\

    \While{$z_t^i \neq g^i$ for any $r^i$} 
    {  
        \tcp{plan for next step}
        UNDE $\gets \mathcal{R}$ \\
        OCCU $\gets \emptyset$ \\
        update the priorities of UNDE \\
        \While{UNDE $\neq \emptyset$} 
        {
            $r^i \gets $ the robot with highest priority in UNDE \\
            \texttt{PBCR}($r^i$, None, None) \\
        }
    }

    \SetKwFunction{FMain}{PBCR}
    \SetKwProg{Fn}{Function}{:}{}
    \Fn{\FMain{$r^i$, $r^j$, $z^j$}}{
        UNDE.$remove(r^i)$\\
        $C \gets $ \texttt{GetValidNeighbors}($p^i_t$)\\
        sort $C$ in decresing order of designed heuristics $H$\\
        \For{$z \in C$}
        {
            $C.remove(z)$\\
            \If{\texttt{Collide}($z$,OCCU$\cup\{z^j_t,z^j\}$)}
            {
                continue\\
            }
            OCCU.$add(z)$\\
            \If{$\exists k \in$ UNDE s.t. \texttt{Collide}($z^k_t,z$)}
            {
                \If{\texttt{PBCR}($r^k$, $r^i$, $z$) is valid}
                {
                    $z^i_{t+1} \gets z$\\
                    \Return valid\\
                }
                \Else{
                    OCCU.$remove(z)$\\
                }
            }
        }
        UNDE.$add(r^i)$\\
        \Return invalid\\
    }
    
\end{algorithm}

Despite achieving state-of-the-art time efficiency, PBCR has several limitations. First, PBCR resolves conflicts by assigning priority order, but this approach lacks sufficient cooperation among robots, often resulting in suboptimal trajectories and lower solution quality. Second, PBCR updates the priority order at each planning step, which can lead to deadlock situations where conflicting robots cannot find a solution. Although careful design of heuristics (\textit{H} in function \textit{PBCR}) can mitigate some deadlock scenarios, duplicate configurations ($\{z_t^1,z_t^2,\dots,z_t^m\}$ at timestep $t$ is defined as a configuration) remain challenging to avoid, ultimately affecting both solution quality and efficiency.

\subsection{ESCoT Algorithm} 
\label{ESCoT}

We transform the representation of the solution $\mathcal{P}$ for the MVTP problem before introducing the ESCoT algorithm. Utilizing the concept of configuration introduced in the previous section, $\mathcal{P}$ can be reconstructed as $\mathcal{P'}$:

\begin{equation*} 
    \label{eqn:config}
    \begin{aligned}
    \mathcal{P} &= \{p^1, \dots, p^m\} = \{[z_0^{1},\dots,z_{T}^{1}], \dots, [z_0^{m}, \dots, z_{T}^{m}]\} \\
    \mathcal{P'} &= \{[z_0^{1},\dots,z_0^{m}], \dots, [z_T^{1}, \dots, z_T^{m}]\} = \{Q_0, \dots, Q_T\}
    \end{aligned}
\end{equation*}

\begin{figure}[htbp]
	\centering
	\includegraphics[width=8.5cm]{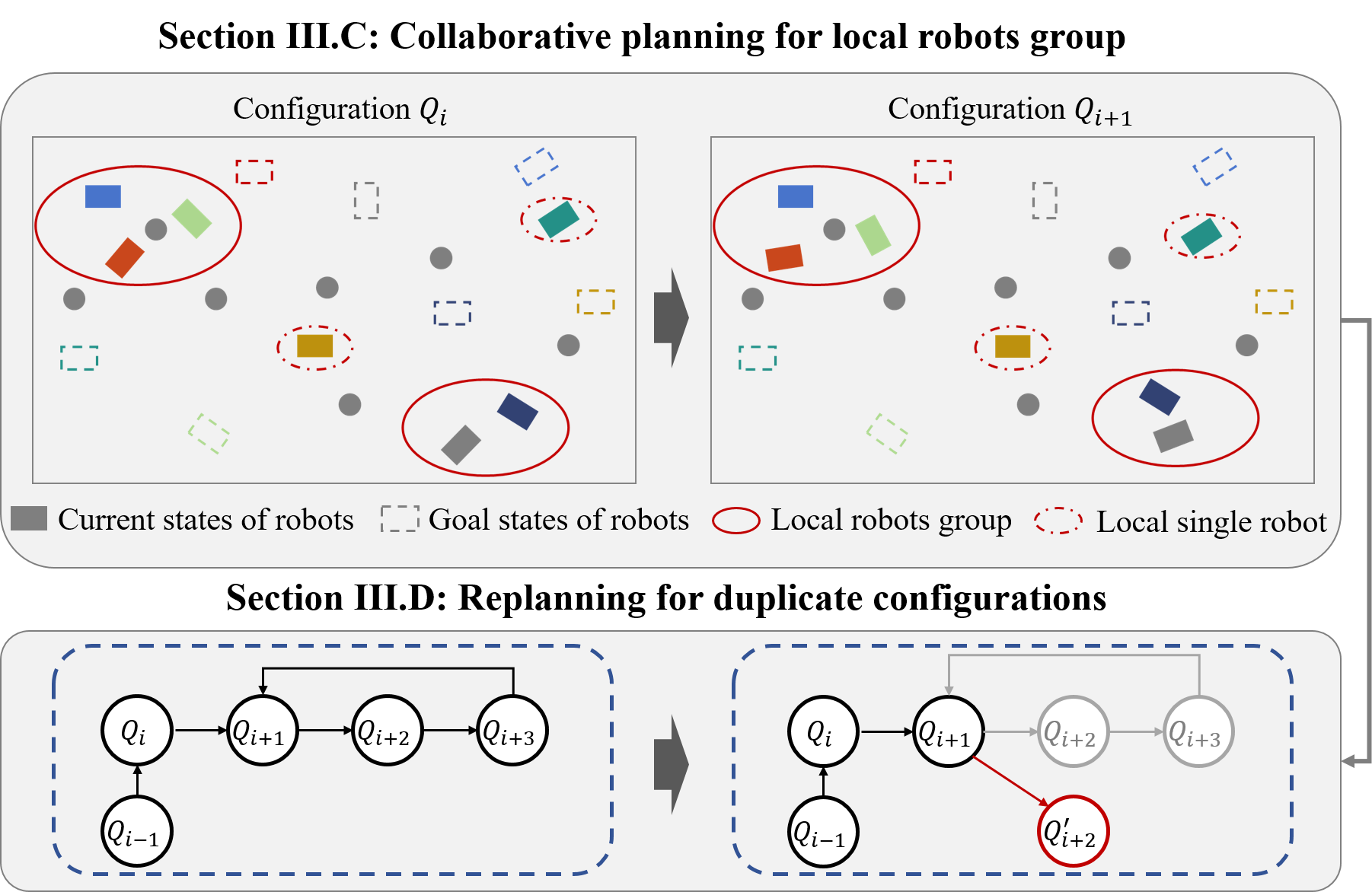}
	\caption{The main process of the ESCoT algorithm, including two key components: collaborative planning for local robots group and replanning for duplicate configuration.}
	\label{fig:escot}
\end{figure}

Now we describe the detailed process of the ESCoT algorithm. The pseudo-code is presented in Algorithm~\ref{algo:escot}. We still adopt the step-based approach, where at each planning step, the next configuration $Q_{\text{new}}$ 
is determined based on the current state of the robots (line 3). However, instead of directly applying PBCR to generate $Q_{\text{new}}$, we introduce a local group collaboration planning strategy (lines 11-17), which will be detailed in Section~\ref{subsec:local-collaborate}. For robots that are not subject to group planning, we employ PBCR to generate their next step's states (lines 18-21). After obtaining $Q_{\text{new}}$, we check for duplicates with previously generated configurations. If any duplicates are found, both the duplicate configuration and all subsequent configurations are removed from $\mathcal{P'}$, and then $Q_{\text{new}}$  is added to it (lines 4-7). This replanning strategy will be discussed in Section~\ref{subsec:replan}. To better illustrate the ESCoT algorithm, we visualize the main process of it in Fig~\ref{fig:escot}.

\begin{algorithm}[hbtp]
    \caption{The ESCoT Algorithm}
    \label{algo:escot}
    \SetKwInOut{Input}{Input}
    \SetKwInOut{Output}{Output}

    \Input{$\mathcal{R},\mathcal{W},\mathcal{O},\mathcal{S},\mathcal{G}$}
    \Output{$\mathcal{P'}= \{Q_0, \dots, Q_T\}$}

    $Q_0 \gets \mathcal{S}$ \\

    \While{$\mathcal{P'}.back \neq \mathcal{G}$} 
    {
        $Q_{\text{new}}$ = \texttt{GenerateConfig}($\mathcal{P'}.back$)\\
        \If{$\exists Q_i \in \mathcal{P'}$ s.t. $Q_i=Q_{\text{new}}$}
        {
            $\mathcal{P'}.remove(Q_i,...,\mathcal{P'}.back)$\\
            \texttt{AddRandomness}($Q_{\text{new}}$)\\
        }
        $\mathcal{P'}.add(Q_{\text{new}})$\\
    }

    \SetKwFunction{FMain}{GenerateConfig}
    \SetKwProg{Fn}{Function}{:}{}
    \Fn{\FMain{$Q_{\text{cur}}$}}{
        UNDE $\gets \mathcal{R}$ \\
        OCCU $\gets \emptyset$ \\
        robots' groups $\Lambda \gets $ \texttt{GroupRobots}($Q_{\text{cur}},\mathcal{R}$)\\
        \For{group $\lambda  \in \Lambda $}
        {
            \If{\texttt{PreCheck}($\lambda$) is true}
            {
                UNDE.$remove(\lambda) $ \\
                trajectories $\Pi=$\texttt{GetOptTraj}($Q_{\text{cur}}[\lambda],\mathcal{G}[\lambda],\mathcal{W},\mathcal{O}$)\\
                $Q_{\text{new}}[\lambda] \gets \Pi_1$\\
                OCCU.$add(\Pi_1)$
            }
        }
        update the priorities of UNDE \\
        \While{UNDE $\neq \emptyset$} 
        {
            $r^i \gets $ the robot with highest priority in UNDE \\
            \texttt{PBCR}($r^i$, None, None) \\
        }
        \Return $Q_{\text{new}}$\\
    }  
\end{algorithm}

% strategy 1:
\subsection{Collaborative Planning for Local Robot Groups}
\label{subsec:local-collaborate}

As shown in Fig.~\ref{fig:escot}, the collaborative planning for local robot groups is primarily used in the function \textit{GenerateConfig}, which generates the next step configuration $Q_{\text{new}}$ based on the current configuration $Q_{\text{cur}}$. First, robots $\mathcal{R}$ are grouped according to some specific rules (function \textit{GroupRobots}), with robots that may conflict within a certain future time window placed in the same group. Then, for each group $\lambda$, a pre-check (function \textit{PreCheck}) is performed to determine if $\lambda$ is suitable to undergo collaborative planning. If the condition is satisfied, the robots within $\lambda$ are planned using a method capable of generating high-quality collaborative trajectories (function \textit{GetOptTraj}), resulting in a set of collision-free trajectories $\Pi$ that connects the current and goal states of the robots in group $\lambda$. For simplicity, in Algorithm~\ref{algo:escot} we use $Q_{\text{cur}}[\lambda]$ to denote the current states of the robots in $\lambda$, and $\mathcal{G}[\lambda]$ to represent their goal states. Finally, $\Pi_1$, the first step of the collaborative trajectories $\Pi$, is assigned to the corresponding positions in $Q_{\text{new}}$.

For robots that are either not grouped or whose group fails the pre-check, the next states is generated using the PBCR method. Note that, in lines 18–21 of Algorithm~\ref{algo:escot}, ESCoT borrows the \textit{PBCR} function in Algorithm~\ref{algo:pbcr}. However, instead of directly assigning $z$ to $Q_{t+1}$ in $\mathcal{P'}$, it is used to fill in the states of $Q_{\text{new}}$ that could not be generated through collaborative planning.

Through this strategy, ESCoT is able to retain the time efficiency of the step-based method while improving the quality of local conflict resolution, thereby enhancing the overall result.

% strategy 2:
\subsection{Replanning for Duplicate Configurations}
\label{subsec:replan}
The design of the replanning strategy primarily aims to address the issue of local optima or deadlock in step-based methods, which is particularly common in dense scenarios. We address this problem through the replanning for duplicate configurations in Algorithm~\ref{algo:escot} (lines 2–7), which is also visualized in Fig.~\ref{fig:escot}. After generating $Q_{\text{new}}$ using the \textit{GenerateConfig} function, ESCoT first checks if $Q_{\text{new}}$ already exists in $\mathcal{P'}$ before adding it. After generating $Q_{\text{new}}$ using the \textit{GenerateConfig} function, ESCoT first checks if $Q_{\text{new}}$ already exists in $\mathcal{P'}$ before adding it. In such cases, ESCoT removes $Q_{i+1}$ and all subsequent configurations from $\mathcal{P'}$ ($Q_{i+1}$ to $Q_{i+3}$ in Fig.~\ref{fig:escot}). Then, the function \textit{AddRandomness} introduces some randomness into $Q_{\text{new}}$, followed by adding it into $\mathcal{P'}$. The purpose of introducing randomness is to avoid planning the same configuration (such as $Q_{i+2}$) in the next step, and instead generate a non-duplicate configuration $Q^{'}_{i+2}$.

Compared to PBCR, the replanning strategy of ESCoT, on the one hand, reduces the occurrence of duplicate configurations, thereby improving the solution quality. On the other hand, the introduction of randomness makes ESCoT more likely to escape from local optima or deadlocks, increasing the chances of finding a feasible solution especially in dense environments.

\subsection{Implementation Detials}
\label{subsec:imple}
\subsubsection{Robot grouping and pre-checking}
The purpose of robots grouping and pre-checking is to identify suitable local robot groups for collaborative planning. In practical applications, we group robots that are likely to experience conflicts in the near future based on their distance and orientation. However, collaborative planning is not applicable to every robot group. If a group contains too many robots, the difficulty of collaborative planning increases significantly. This is one of the primary reasons why many MVTP solvers struggle to achieve high success rates in dense scenarios. In such cases, local grouping loses its effectiveness. ESCoT addresses this challenge by applying pre-checking to robot groups, which helps avoid local MVTP problems with unacceptable complexity. Currently, this pre-checking is implemented by constraining the number of robots in each group, ensuring that collaborative planning is only applied to groups with a moderate conflict resolution complexity, thereby improving the feasibility of solving local conflicts.

\subsubsection{Collaborative planning solvers}
The collaborative planning solvers used in ESCoT need to provide high-quality solutions for smaller-scale MVTP problems, such as CL-CBS \cite{wen2022cl}, CSDO \cite{yang2024csdo}, etc. In this paper, we adopt the ECCR algorithm proposed in \cite{guo2024decentralized} as the collaborative planning solver for ESCoT. ECCR enhances the CL-CBS algorithm by replacing the traditional low-level spatiotemporal hybrid A* planner with a focal hybrid A* search \cite{barer2014suboptimal}. ECCR optimizes CL-CBS in the same way that ECBS optimizes CBS, by significantly reducing the number of high-level expansions, thus enhancing the algorithm's efficiency.

\subsubsection{Adding randomness to configurations}
As previously mentioned, when a duplicate configuration is encountered, we introduce randomness to increase the probability of generating a new configuration. This can be achieved through various techniques. For instance, one is to randomly assign certain robots to take specific actions in the next step, rather than computing them algorithmically. Alternatively, the actions taken by some robots to reach their current states can be modified randomly, thereby these robots are encouraged to choose different actions in the next step. Both of these techniques are employed in ESCoT.

\section{Experiments and Validation}
\label{Section 4}
In this section, we first introduce the experimental settings, the benchmark used, and the baselines for comparison. Then, we conduct experiments in typical conflict scenarios, sparse environments, and dense environments to evaluate the performance of the ESCoT algorithm. Finally, the ESCoT algorithm is deployed on practical robots to demonstrate its applicability in real-world scenarios.

\subsection{Settings, Benchmark and Baselines}
For each instance in the experiments, a time limit of \textbf{one minute} is set. If the algorithm fails to produce a solution within this time frame, it is considered a failure. Due to the non-deterministic nature of both ESCoT and PBCR, these methods are run five times for the same instance, with each run subject to a 0.2-minute time limit. All the algorithms tested are implemented in C++, and all experiments are conducted on a computer equipped with an Intel Core i7-12700KF CPU, an NVIDIA GeForce RTX 3080 GPU, and 32GB of RAM, running on a Linux system.

\begin{figure}[htbp]
	\centering
	\includegraphics[width=8.5cm]{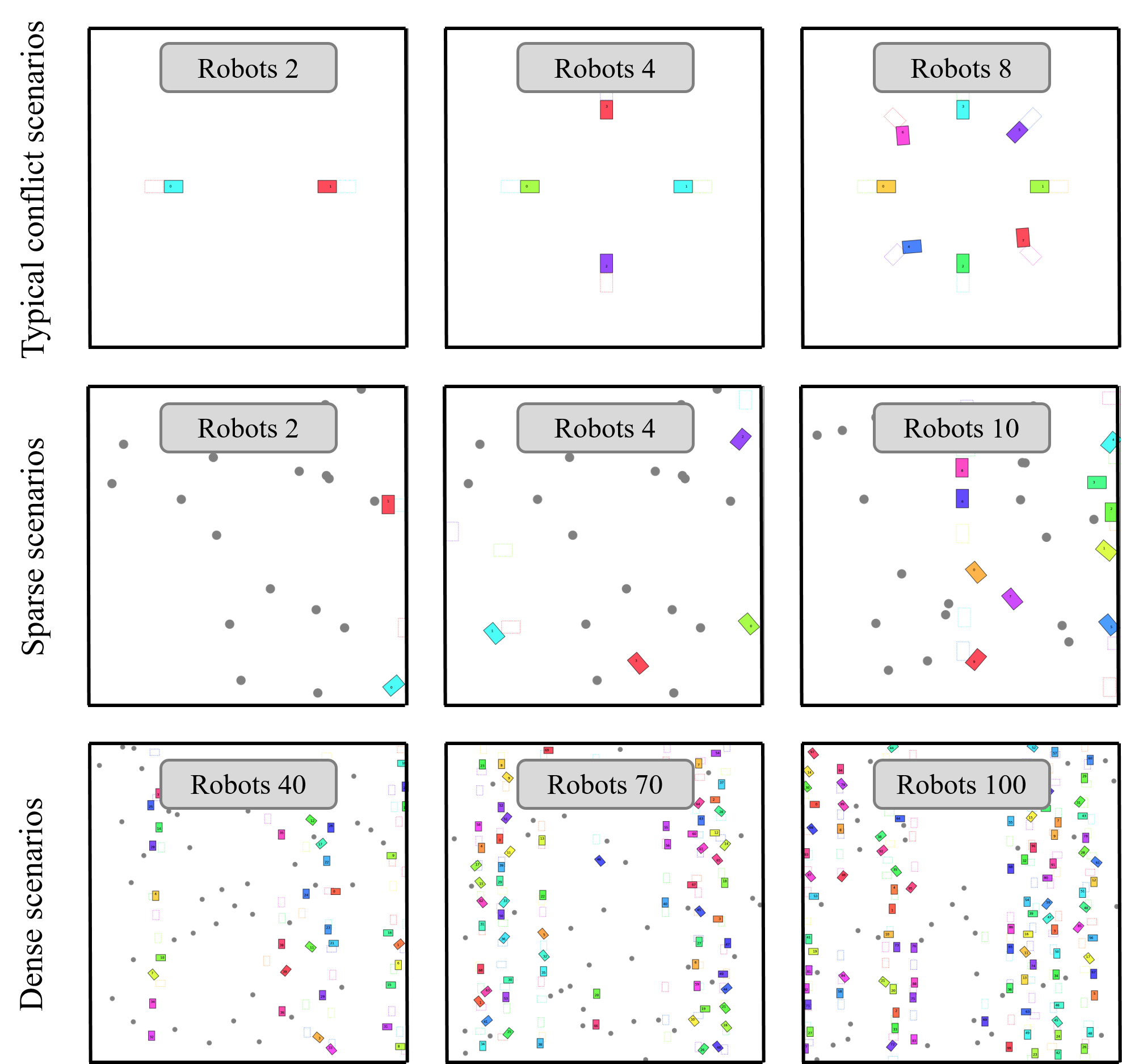}
	\caption{Example instances in the benchmark from typical conflict scenarios, sparse scenarios and dense scenarios.}  
	\label{fig:benchmark}
\end{figure}

The benchmark in this paper consists of three scenarios: typical conflict scenarios, sparse scenarios, and dense scenarios. In the typical conflict scenario, we have different numbers of car-like robots driving toward each other. In the sparse scenario, the map is $50m\times50m$, with randomly distributed obstacles, and the number of robots ranges from 2 to 10. For each robot count, there are 60 instances, so the sparse scenario in the benchmark includes a total of 300 instances. The dense scenario includes different types of maps (with or without obstacles, denoted as "empty" and "random"), different map sizes ($50m\times50m$ and $100m\times100m$), and varying numbers of robots (5 to 120). For each combination of map type, map size, and robot count, there are also 60 instances, so the dense scenario contains a total of 1,740 instances in the benchmark. For the sparse and dense scenarios in the benchmark, the initial and goal states of the robots, as well as the positions of the obstacles, are randomly generated. In the simulation, we consider all car-like robots are homogeneous and share the following parameters: $l=3$m, $w=2$m, $l_b=2$m, $v_{\text{max}}=1$m/s, $\phi_{\text{max}}=40.1$°, $\Delta t=1$s. Fig.~\ref{fig:benchmark} illustrates several instances from various scenarios in the benchmark.

As for the baselines, our primary comparison algorithm is PBCR, which is also a step-based method and achieves scalable SOTA on certain metrics of the MVTP problem \cite{guo2024decentralized}. In addition, we have selected three representative methods: the centralized planning method CL-CBS \cite{wen2022cl}, ECCR \cite{guo2024decentralized}, and the decentralized planning method SHA* \cite{wen2022cl}, as baselines.

\subsection{Typical Conflict Scenarios}
\begin{figure}[htbp]
	\centering
	\includegraphics[width=8.5cm]{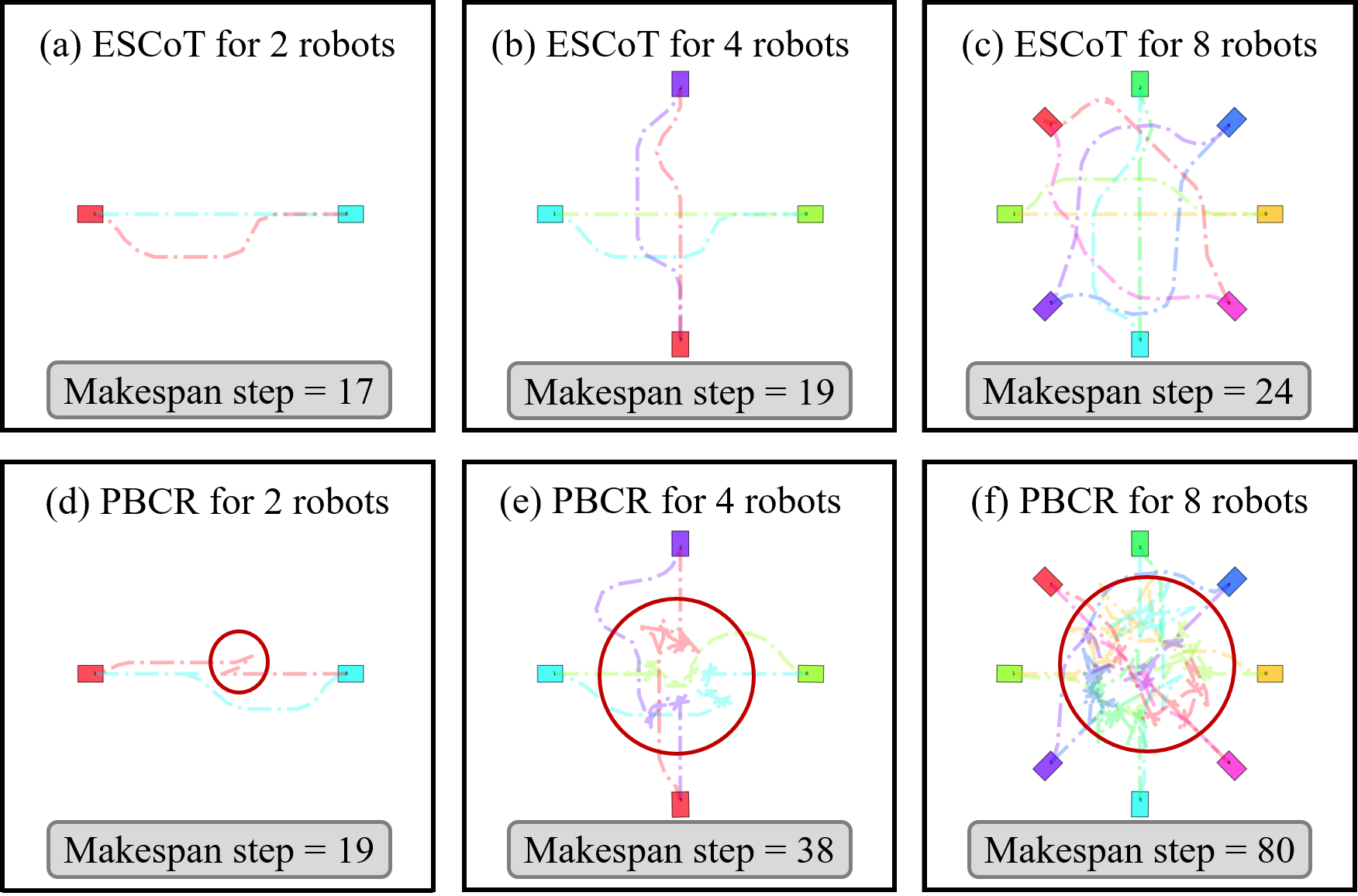}
	\caption{Planning results when use ESCoT and PBCR to solve the MVTP problem in the three typical conflict scenarios. In (d) ,(e) and (f), the red circles highlight the local 'oscillation' observed in the PBCR algorithm's results, where robots experience oscillatory behavior while attempting to resolve conflicts.}
	\label{fig:typi-scen}
\end{figure}

We designed three typical conflict scenarios, as shown in the first row of Fig.~\ref{fig:benchmark}, where different numbers of robots need to drive toward each other to reach their respective goal states. For non-step-based methods, solving these MVTP problems is not difficult, as the problem size is relatively small (also a sparse scenario), and high-quality solutions can be obtained within a short time. However, for step-based methods like PBCR, due to its nature of planning for the next step at each iteration, it is difficult for the solver to foresee conflicts that may arise in the near future (even after just two steps). As a result, robots often experience multiple rounds of local 'oscillation' to resolve conflicts. Although feasible solutions can still be found, the solution quality is significantly reduced. We use both ESCoT and PBCR to solve the MVTP problem in the three typical conflict scenarios, with the resulting trajectories shown in Fig.~\ref{fig:typi-scen}.

% 对fig 5做一下解释。
It can be seen that although ESCoT is a step-based method, the collaborative planning strategy for local robot groups significantly improves the quality of local conflict resolution. Furthermore, as the complexity of the local conflicts increase, the quality improvement becomes more pronounced. In the case of a two-robot conflict scenario, ESCoT reduced the makespan step by approximately 10\% compared to PBCR. This reduction increased to 50\% and 70\% in the scenarios with four and eight robots, respectively.

\subsection{Sparse Scenarios}

To further illustrate the improvement of solution quality achieved by ESCoT in sparse scenarios, we conducted experiments on the randomly generated sparse scenarios of the benchmark. We compared the performance of all algorithms across three aspects: 1) \textbf{Success rate}, which is the proportion of instances that were solved successfully within the time limit for each experimental configuration (robot count, map size, and map type). 2) \textbf{Makespan step}, which refers to the number of planning steps required for the last robot to reach its goal state and remain stationary. We averaged the makespan steps for successfully solved instances, to obtain the algorithm's result for that configuration. 3) \textbf{Runtime}, which measures the time taken by the algorithm to solve the problem. For unsolved instances, we assume the runtime equals the time limit and average this across all instances.

\begin{figure}[htbp]
	\centering
	\includegraphics[width=1\linewidth]{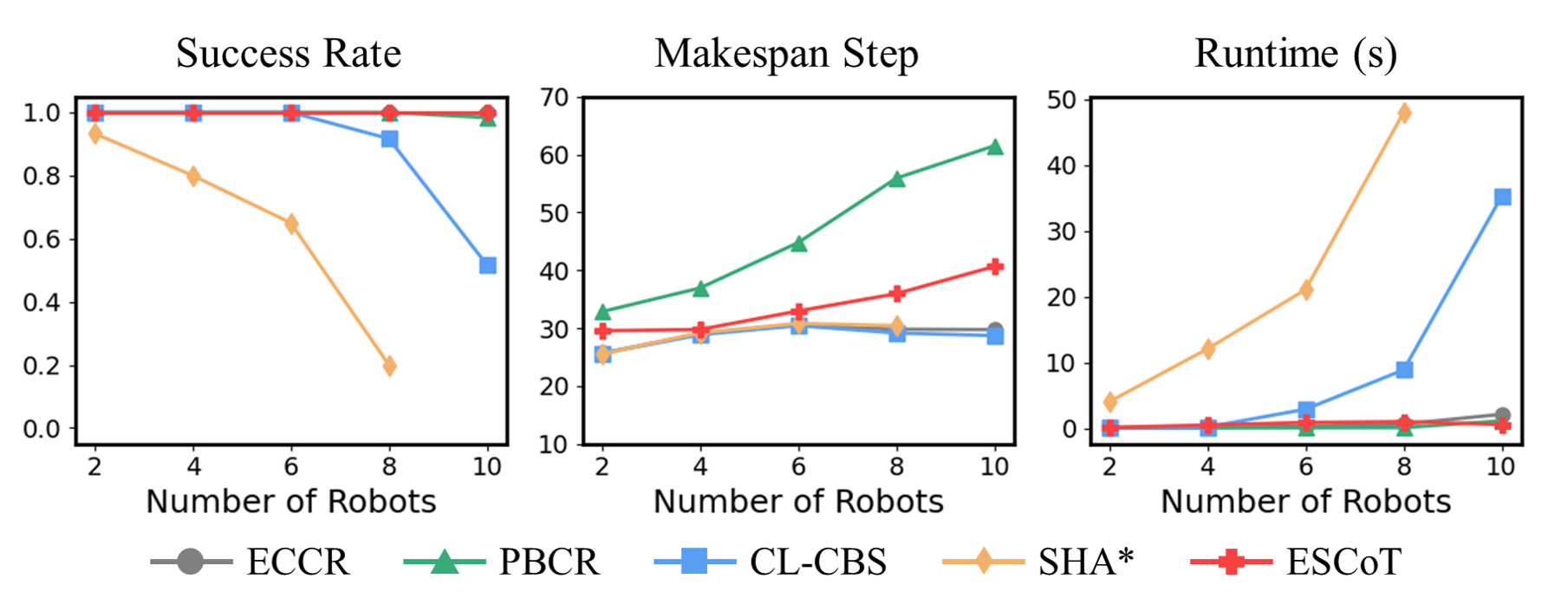}
	\caption{Simulation results for sparse scenarios on a 50*50 random map.}
	\label{fig:com-sparse}
\end{figure}

Detailed evaluation results are presented in Fig.~\ref{fig:com-sparse}. It can be observed that in sparse scenarios, ESCoT maintains the high solution efficiency characteristic of step-based methods, outperforming others in both success rate and runtime. In terms of makespan step, ESCoT significantly reduces the number of steps compared to PBCR (for instance, with 10 robots, the reduction is 34\%), narrowing the gap in solution quality between step-based methods and centralized planning methods. Therefore, it can be concluded that ESCoT, in sparse scenarios, improves the solution quality of step-based methods while maintaining high solving efficiency.

\subsection{Dense Scenarios}

\begin{figure}[tbhp]
	\centering
	\includegraphics[width=1\linewidth]{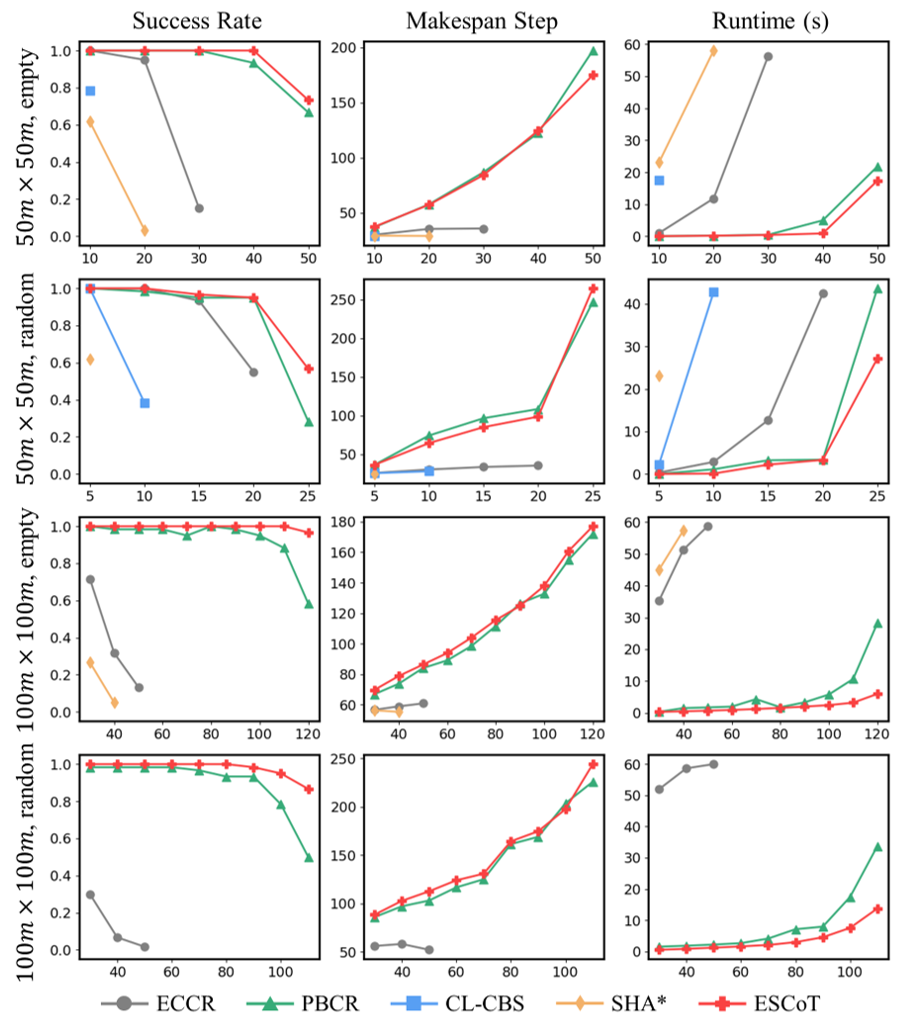}
	\caption{Simulation results for dense scenarios on maps with different types and sizes. To make the figures layout more compact, we omit the x-axis labels for each figure, which is the number of robots.}
	\label{fig:com-dense}
\end{figure}

We then evaluate the performance of ESCoT against the baselines on dense scenarios. Fig.~\ref{fig:com-dense} presents the results for all map configurations. In dense scenarios, due to the high complexity of the MVTP problem, step-based methods like PBCR and ESCoT demonstrate significant advantages in terms of success rate and runtime. CL-CBS, as a centralized planning method, can only handle instances with fewer than 10 robots on a $50m\times50m$ map. When the map size increases or the number of robots grows, CL-CBS fails to find a solution within the time limit. ECCR, an enhanced algorithm over CL-CBS, outperforms CL-CBS by solving more instances in less runtime with almost no loss in solution quality. However, as the MVTP problem size increases, ECCR's performance also declines rapidly. Specifically, on a $50m\times50m$ map with more than 30 robots or a $100m\times100m$ map with more than 50 robots, ECCR fails entirely. SHA* performs slightly worse than ECCR, solving fewer instances.

In our simulations, ESCoT achieves the best success rate and runtime performance across both empty and random maps. Even with the maximum number of robots for each map configuration, ESCoT maintains a success rate of over 50\%, significantly outperforming PBCR and other baseline algorithms. Regarding runtime, ESCoT retains excellent time efficiency due to the inherent characteristics of step-based methods. In terms of makespan step, ESCoT has not yet reached the solution quality of centralized planning methods, partly because centralized methods often struggle to find solutions for more difficult MVTP instances, whereas ESCoT succeeds by resolving conflicts locally. Nevertheless, for easier configurations (e.g., 10 robots on a $50m\times50m$ empty map, or 5 robots on a $50m\times50m$ random map), the performance gap between ESCoT and centralized methods is minimal. Compared to PBCR, ESCoT's success rate is further improved, benefiting from its replanning strategy for duplicate configurations. The improvement becoming more pronounced as the problem difficulty increases.

Based on the results, ESCoT outperforms all baseline methods in dense scenarios, further expanding the performance boundaries of step-based methods for solving MVTP problems.

\subsection{Practical Robot Tests}
In this section, we utilized a team of small physical robots to validate the ESCoT algorithm. The experimental platform used consists of toio robots (https://toio.io/), which communicate with the control center via Bluetooth and navigate on a specialized mat by continuously receiving coordinate instructions.

\begin{figure*}[bthp]
    \begin{center}    
         \includegraphics[width=0.95\linewidth]{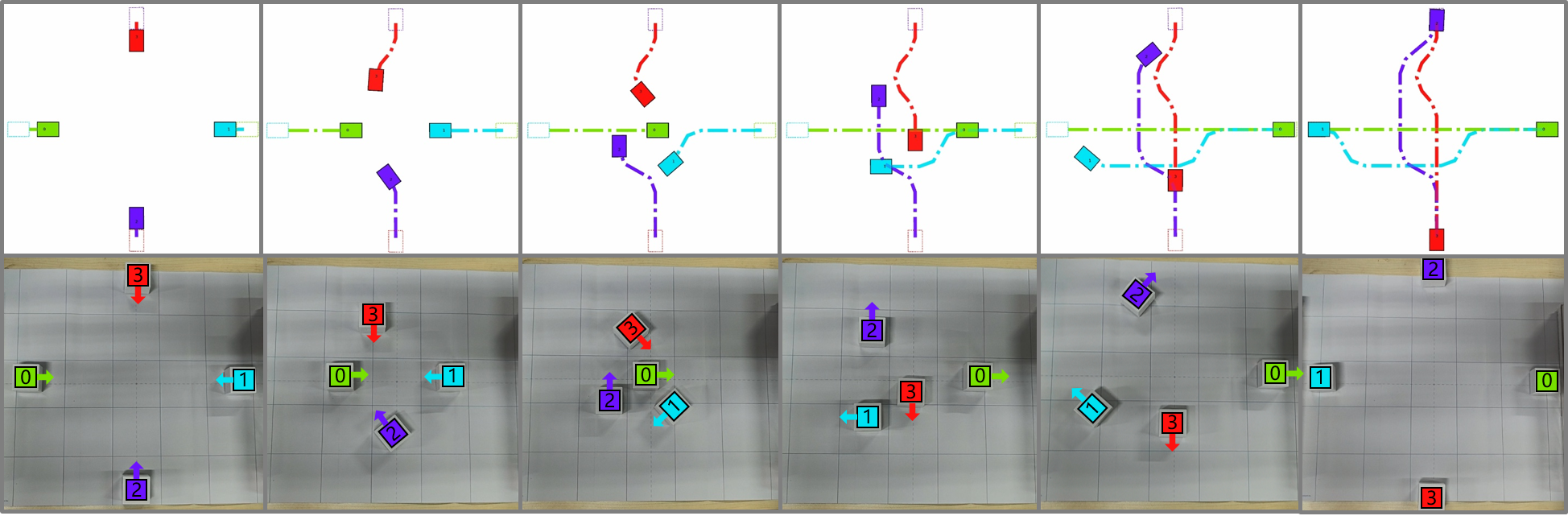}
    \end{center}
   
    \caption{The snapshots of the practical robot test demo. Rectangular boxes in different colors represent the robots, while the corresponding dashed rectangles indicate their respective goal states. To enhance clarity, arrows are added to the practical robots, illustrating their yaw orientation at each time step.}
    \label{fig:real-robot}
\end{figure*}

We used the typical conflict scenario involving four robots (Fig.~\ref{fig:typi-scen}(b)) as the practical robot test scenario. To facilitate real-world testing, we implemented a coordinate transformation between the dedicated mat and the simulation environment. The predefined start and goal configurations were transmitted to the control center (a laptop here) where the ESCoT algorithm was executed to generate planned trajectories for each robot. The computed trajectories were then communicated to the robots via Bluetooth for execution. Fig.~\ref{fig:real-robot} presents a series of snapshots illustrating this scenario, comparing the simulation result with real-world robot experiment. The experimental implementation was developed in Python. The results demonstrate that the robots successfully followed their designated trajectories, confirming the applicability of the ESCoT algorithm in practical robotic scenarios.

\section{Conclusions}
\label{Section 5}
This paper introduces ESCoT, an enhanced step-based coordinate trajectory planning method for multiple car-like robots. We equip ESCoT with two key strategies, collaborative planning for local robot groups and replanning for duplicate configurations, which are effective in improving the performance of step-based methods for MVTP. Through an extensive set of experiments, we demonstrate that ESCoT improves the solution quality of step-based methods in sparse scenarios (up to 70\% for typical conflict scenarios and 34\% for randomly generated scenarios), while maintaining high solving efficiency. In dense scenarios, ESCoT outperforms all baselines, further expanding the performance boundaries of step-based methods. Finally, practical robot tests are conducted to verify the applicability of the algorithm in real-world robotic scenarios. In the future, we will continue to explore the potential of step-based methods for MVTP, aiming to achieve state-of-the-art solution quality while maintaining high solving efficiency.

% This paper proposes a generalized control revision method for autonomous driving safety. Based on a perception data conversion layer and unified traffic element constraint representation, the control revision module can ensure autonomous vehicle control safety comprehensively. The proposed method was tested on multiple simulators including CARLA, SUMO, and OnSite and validated on a real-world platform MCCT. Experiments proved that our method could revise the danger behaviors and avoid accidents with various kinds of perception data and planning backbone under different road topologies. Tests on continuous traffic flow scenes also showed that the accident rate can be greatly reduced by applying control revision. Validations on real-world platforms verified the feasibility.

\addtolength{\textheight}{0cm}   % This command serves to balance the column lengths

%%%%%%%%%%%%%%%%%%%%%%%%%%%%%%%%%%%%%%%%%%%%%%%%%%%%%%%%%%%%%%%%%%%%%%%%%%%%%%%%

%%%%%%%%%%%%%%%%%%%%%%%%%%%%%%%%%%%%%%%%%%%%%%%%%%%%%%%%%%%%%%%%%%%%%%%%%%%%%%%%



% \section*{ACKNOWLEDGMENT}

% The preferred spelling of the word ÒacknowledgmentÓ in America is without an ÒeÓ after the ÒgÓ. Avoid the stilted expression, ÒOne of us (R. B. G.) thanks . . .Ó  Instead, try ÒR. B. G. thanksÓ. Put sponsor acknowledgments in the unnumbered footnote on the first page.

\bibliography{IEEEabrv, mybibfile}

\end{document}